\pdfoutput=1

\documentclass[11pt]{article}

\usepackage[preprint]{acl}

\usepackage{times}
\usepackage{latexsym}

\usepackage[T1]{fontenc}

\usepackage[utf8]{inputenc}

\usepackage{microtype}

\usepackage{inconsolata}

\usepackage{graphicx}
\usepackage{booktabs}
\usepackage{makecell}
\usepackage{float}

\usepackage{amssymb}

\usepackage{amsmath}
\usepackage{amsthm}
\usepackage{algorithm}
\usepackage{algpseudocode}

\usepackage{cleveref}
\usepackage{multirow}
\usepackage{multicol}
\usepackage{arydshln}
\usepackage{mathrsfs}
\usepackage{enumitem}
\newcommand{\exper}[1]{\textsc{#1}}

\DeclareMathSymbol{\minus}{\mathbin}{AMSa}{"39}

\usepackage{tcolorbox}
\tcbset{
  mybox/.style={
    colback=gray!10!white,            
    colframe=gray!60!white,           
    coltitle=white,                   
    fonttitle=\bfseries,              
    colbacktitle=gray!60!white,        
    title filled,                     
    coltitle=white,                   
    boxrule=1pt,                      
    rounded corners=all,              
    toptitle=2pt,
    bottomtitle=2pt,
    left=10pt,            
    right=10pt,           
    top=5pt,             
    bottom=5pt,           
  }
}
\newtcolorbox[auto counter, number within=section]{template}[2][]{
  title={Prompt Template: #2}, 
  mybox,
  #1
}

\newtcolorbox[auto counter, number within=section]{format}[2][]{
  title={}, 
  mybox,
  #1
}

\renewcommand{\algorithmiccomment}[1]{\bgroup\hfill\textcolor{gray}{//~#1\egroup}}
\title{Length Controlled Generation for Black-box LLMs}

\author{\textbf{Yuxuan Gu}$^{1}$ \quad \textbf{Wenjie Wang}$^{2}$ \quad \textbf{Xiaocheng Feng}$^{1 3}$ \textbf{Weihong Zhong}$^{1}$\\ \quad \textbf{Kun Zhu}$^{1}$ \quad \textbf{Lei Huang}$^{1}$ \quad \textbf{Tat-Seng Chua}$^{2}$ \quad \textbf{Bing Qin}$^{1 3}$\\
$^{1}$Harbin Institute of Technology \quad $^2$ National University of Singapore \quad  $^3$ Peng Cheng Laboratory\\
\texttt{\{yxgu,xcfeng,whzhong,kzhu,lhuang,qinb\}@ir.hit.edu.cn}\\
\texttt{wenjiewang96@gmail.com \quad dcscts@nus.edu.sg}\\}

\begin{document}
\maketitle
\begin{abstract}
Large language models (\exper{Llm}s) have demonstrated impressive instruction following capabilities, while still struggling to accurately manage the length of the generated text, which is a fundamental requirement in many real-world applications. Existing length control methods involve fine-tuning the parameters of \exper{Llm}s, which is inefficient and suboptimal for practical use. 
In this paper, we propose a novel iterative sampling framework for text length control, integrating the Metropolis-Hastings algorithm with an importance sampling acceleration strategy. 
This framework efficiently and reliably regulates \exper{Llm}s to generate length-constrained text without modifying the underlying parameters, thereby preserving the original capabilities of \exper{Llm}s.
Experimental results demonstrate that our framework achieves almost 100\% success rates of length control on \exper{Llama}3.1 for tasks such as length-controlled abstractive summarization and length-constrained instruction following, with minimal additional computational overhead. 
This also highlights the significant potential of our method for precise length control across a broader range of applications, without compromising the versatility of \exper{Llm}s.
\end{abstract}

\section{Introduction}
Recent advancement of pre-trained large language models (\exper{Llm}s) has significantly improved the performance of various natural language processing tasks \cite{vaswani2017attention, devlin2018bert, brown2020language}.
\exper{Llm}s such as \exper{Gpt}-4 \cite{achiam2023gpt} and \exper{Llama} \cite{touvron2023llama, touvron2023llama2, dubey2024llama} exhibit exceptional capabilities to follow instructions \cite{ouyang2022training}, allowing them to generate text aligning closely with user intentions.
Applications such as dialogue generation \cite{yi2024survey}, code completion \cite{jiang2024survey}, and reasoning \cite{plaat2024reasoning} have benefited greatly from these advances, establishing \exper{Llm}s as the core component in building general AI systems.

Despite the strong generative capability, \exper{Llm}s still struggle to precisely manage the length of generated text \cite{wang-etal-2024-positionid, huang-etal-2024-decoding, li-etal-2024-ruler}, due to inherent architectural limitations such as subword tokenization \cite{sennrich2015neural, devlin2018bert} and autoregressive decoding \cite{sutskever2014sequence, vaswani2017attention, brown2020language}. 
This issue is critical because length control is a fundamental requirement in many real-world applications. 
For example, summarization tasks often require outputs of specific lengths to balance informativeness and conciseness \cite{fan2017controllable, liu2018controlling, liu2022length, jie-etal-2024-prompt}.
In addition, \exper{Llm}-based chatbots favor longer responses due to the length bias introduced in pairwise preference optimization \cite{singhal2023long}, which undermines the fairness of model evaluation \cite{dubois2024length, yuan2024following} and degrades the user experience in practical conversations.

To address the issue of length control, various methods have been proposed, including fine-tuning based on specifically designed length instructions \cite{yuan2024following, wang-etal-2024-positionid,li-etal-2024-ruler} and reinforcement learning with length feedback \cite{stiennon2020learning, jie-etal-2024-prompt}. 
However, we argue that it is necessary to design length control methods tailored for black-box \exper{Llm}s for the following reasons:
(1) Fine-tuning \exper{Llm}s specifically for length control requires extensive computational resources and can degrade their general-purpose utility \cite{lin-etal-2024-mitigating}. Worse still, not all \exper{Llm}s are open source. The fine-tuning methods cannot be applied to black-box \exper{Llm}s. 
(2) 
Length control has been actually considered in the instruction tuning phase of \exper{Llm}s \cite{wang2022self, alpaca}. 
As such, a superior and more efficient solution is to activate the inherent length-following capabilities within \exper{Llm}s rather than undertaking a costly retraining process. 

We propose a novel framework for black-box \exper{Llm}s that operates length control without the need for parameter training. 
The length controlled generation can be viewed as sampling from a target distribution, which is influenced simultaneously by the length constraint and language probability. 
However, it is intractable to directly sample from this distribution, and we utilize an iterative sampling framework called Metropolis-Hastings \cite{metropolis1953equation, hastings70}, which is a classic and prevalent Markov chain Monte Carlo (MCMC) method specially suited for this complex scenario.
In detail, our framework initiates from the original output of \exper{Llm} and iteratively produces candidate outputs conditioned on the previous ones via a proposal distribution. The acceptance or rejection of these candidates is determined by their comparative advantage over previous candidates, which is quantified as an acceptance distribution that involves: the alignment with the target length, the generative probability density of the \exper{Llm}, and the probability density of the proposal distribution.
Furthermore, we leverage importance sampling \cite{kahn1953methods, owen2000safe} in the proposal distribution to accelerate the iteration process, where candidates with lengths closer to the desired target are more likely to be sampled. We treat the \exper{Llm} as an immutable component, enabling the integration of effective length control mechanisms across the broadest possible spectrum of \exper{Llm}s.

We assess the effectiveness of our method on several tasks, including the abstract text summarization task with precise length control and the instruction following task with maximum length constraint.
Experimental results demonstrate that our black-box approach significantly improves existing \exper{Llm}s in length control and achieves the state-of-the-art performance without compromising the quality of generated contents.
Specifically, in the case of the \exper{Llama}3.1 model \cite{dubey2024llama}, our method achieves success rates close to $100\%$ of the length control in only five iterations at most, highlighting its efficiency and practicality.
Our contributions are summarized as follows:
\begin{enumerate}[itemsep=0pt, parsep=0pt, topsep=0pt]
    \item We propose a novel framework for black-box \exper{Llm}s, offering more flexible and general length control compared to existing methods.
    \item We introduce an innovative integration of the classic Metropolis-Hastings algorithm with modern \exper{Llm}s, thereby enhancing the efficiency and precision of length control.
    \item 
    We achieve remarkable length control performance in advanced \exper{Llm}s, showcasing the robustness and effectiveness of our framework.
\end{enumerate}
\section{Related Work}
\subsection{Instruction Following}
\exper{Llm}s are endowed with powerful instruction following capabilities in the supervised fine-tuning stage \cite{ouyang2022training,zhou2024lima}.
Despite being able to understand human instructions and handle a broad spectrum of tasks, \exper{Llm}s still leave a large room for improvement in their instruction following capabilities \cite{liu2023makes}.
In addition to training stronger instruction following capabilities \cite{rafailov2024direct,rafailov2024r}, it is also important to better utilize and activate the power of \exper{Llm}s \cite{wei2022chain,yao2023react}.

\begin{figure*}[ht]
    \centering
    \includegraphics[width=1.0\textwidth]{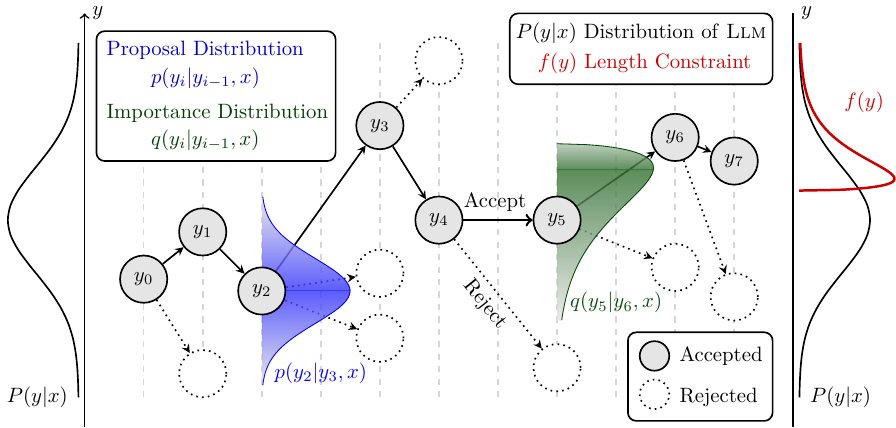}
    \caption{The overall sampling process of our Metropolis-Hastings framework. The iteration starts by sampling an initial state from the distribution of \exper{Llm} $y_0\sim P(y|x)$, and ends at $y_7$, which maximizes the target combination of length constraints and probability densities $\pi(y|x)\propto f(y)P(y|x)$. During each iteration, a new candidate content $y_i$ is generated based on the previous one $y_{i-1}$ via the proposal distribution $p(y_i|y_{i-1},x)$. The generated candidate $y_i$ will be either accepted or rejected considering the degree to which the target objectives are satisfied. 
    We enhance the original proposal distribution by incorporating length constraints, yielding the importance distribution $q(y_i|y_{i-1},x)$, 
    which increases the acceptance rate of candidates and significantly improves the iteration efficiency.}
    \label{fig:mh}
\end{figure*}

\subsection{Length Control}

Controlling the output length is a crucial skill in text generation, particularly for tasks where lengths vary significantly.
Early length controllable generation methods focus on the abstractive summarization task. 
For example, some methods discretize lengths into bins with specialized tokens \cite{fan2017controllable}, introduce length constraint factors to convolutional blocks \cite{liu2018controlling}, or optimize output quality through minimum risk training \cite{makino2019global}.
In addition, length control signals can be incorporated in positional encodings \cite{takase2019positional}, attention units \cite{yu2021lenatten,liu2022length}, and natural language instructions \cite{yuan2024following,wang-etal-2024-positionid,jie-etal-2024-prompt,li-etal-2024-ruler}.
These methods require the length training, which is inefficient when applied to \exper{Llm}s and has the potential to damage general abilities. In contrast, our framework controls the generated length during the inference stage.


\section{Methodology}
\subsection{Overall Framework: Metropolis-Hastings}
As illustrated in \Cref{fig:mh}, we introduce how to apply the Metropolis-Hastings framework \cite{metropolis1953equation, hastings70} to the length control scenario. Given the probability distribution of \exper{Llm}s $P(y|x)$ and the score of length constraint $f(y)$, our target distribution $\pi(y|x)$ is derived as:
\begin{equation}
\pi(y|x) = \frac{f(y)P(y|x)}{\int f(y)P(y|x)\mathrm{d}y},
\end{equation}
where $x$ is the human instruction and $y$ is the response of the target \exper{Llm}.
We cannot directly sample $y$ from the target distribution $\pi(y|x)$ because: (1) $f(y)$ is a deterministic function designed to evaluate length constraints, rather than a probability distribution, which is not suitable for sampling; and (2) the integral of the normalization constant $Z = \int f(y)P(y|x)\mathrm{d}y$ is intractable.

The Markov chain Monte Carlo algorithms can handle the problem by starting from an initial state $y_0\sim P(y|x)$, iteratively generating a collection of states $[y_1,\dots,y_n]$ with a transition distribution $P(y_i|y_{i \minus 1},x)$, and approaching the target distribution $\pi(y|x) = \lim\limits_{n\to \infty}  P(y_0|x) \prod\limits_{i=1}^n P(y_i|y_{i \minus 1},x)$. Therefore, $y_n$ can be considered as sampled from the target distribution $\pi(y|x)$ when $n\to \infty$. 

The Metropolis-Hastings algorithm designs the transition probability as a combination of two steps:
\begin{equation}
    \label{eq:transition}
P(y_i|y_{i \minus 1},x)=p(y_i|y_{i \minus 1},x)\mathcal{A}(y_{i \minus 1}\rightarrow y_i),
\end{equation}
where $p(y_i|y_{i \minus 1},x)$ is the proposal distribution that generates a new candidate $y_i$ given the previous one $y_{i-1}$. The acceptance distribution $\mathcal{A}(y_{i \minus 1}\!\rightarrow\! y_i)$ provides the probability of accepting the proposed candidate $y_i$.
To ensure convergence, $\pi(y|x)$ must be the unique stationary distribution of the Markov chain. Thus the Metropolis-Hastings algorithm further requires the transition probability $P(y_i|y_{i \minus 1},x)$ to fulfill the detailed balance condition, which is a sufficient condition for the stationary distribution,
\begin{equation}
    \label{eq:balance}
    \pi(y_{i \minus 1}|x)P(y_i|y_{i \minus 1},x) = \pi(y_i|x)P(y_{i \minus 1}|y_i,x).
\end{equation}
Based on \cref{eq:transition,eq:balance}, the acceptance distribution is derived to satisfy the following constraint:
\begin{equation}
    \label{eq:A_constraint}
    \begin{aligned}
    &\frac{\mathcal{A}(y_{i \minus 1}\rightarrow y_i)}{\mathcal{A}(y_{i}\rightarrow y_{i \minus 1})} = \frac{\pi(y_{i}|x)p(y_{i \minus 1}|y_i,x)}{\pi(y_{i \minus 1}|x)p(y_{i}|y_{i \minus 1},x)}\\
    &\qquad =\frac{f(y_i)P(y_i|x)p(y_{i \minus 1}|y_i,x)}{f(y_{i \minus 1})P(y_{i \minus 1}|x)p(y_i|y_{i \minus 1},x)},
    \end{aligned}
\end{equation}
where the normalization constant $Z$ cancels, making subsequent calculations convenient.
In addition, the most popular choice of $\mathcal{A}(y_{i \minus 1}\!\rightarrow\! y_i)$ in Metropolis-Hastings that satisfies \cref{eq:A_constraint} is:
\begin{equation}
    \label{eq:A}
    \min\left(1, \frac{f(y_i)P(y_i|x)p(y_{i \minus 1}|y_i,x)}{f(y_{i \minus 1})P(y_{i \minus 1}|x)p(y_i|y_{i \minus 1},x)}\right).
\end{equation}

\begin{algorithm}[t]
\caption{Metropolis-Hastings Algorithm}
\label{alg:framework}
\begin{algorithmic}[1]
\State \textbf{Initialize} the start state $y_0\sim P(y|x)$
\For{$i = 0$ to $n$}
    \State \textit{Propose}: $y_i\sim p(y_i|y_{i \minus 1},x)$ \label{alg:propose}
    \State \textit{Calculate}: $\mathcal{A}(y_{i \minus 1}\!\rightarrow\! y_i)$ \Comment{\cref{eq:A}}
    \State \textit{Randomize}: $u\sim \mathcal{U}(0, 1)$ 
    \If{$u>\mathcal{A}(y_{i \minus 1}\!\rightarrow\! y_i)$} 
        \State $y_i = y_{i \minus 1}$ \Comment{Reject}
    \EndIf \Comment{else Accept}
\EndFor
\State \textbf{Return} $y_n$
\end{algorithmic}
\end{algorithm}

The sampling process of Metropolis-Hastings is illustrated in \cref{alg:framework}. During each iteration loop, a new candidate $y_i$ is generated from the previous one $y_{i \minus 1}$. Whether to accept or reject the new candidate is determined by the acceptance distribution $\mathcal{A}(y_{i \minus 1}\!\rightarrow\! y_i)$, where the randomness is achieved with a uniform distribution $u\sim\mathcal{U}(0,1)$. 

In the black-box setting where direct access to the internal probability outputs of \exper{Llm} is not available, the following key components of the Metropolis-Hastings algorithm require careful and specialized designs: (1) the length constraint score $f(y)$ (\S \ref{sec:length}), which serves as a quantitative metric to assess the degree to which the generated samples adhere to predefined length requirements; 
(2) the probability distribution of \exper{Llm} $P(y|x)$ (\S \ref{sec:target}), which estimates the likelihood of the \exper{Llm} generating the specific sample $y$;
and (3) the proposal distribution $p(y_i|y_{i \minus 1},x)$ (\S \ref{sec:proposal}), which requires tailored construction to effectively generate candidate samples and efficiently explore the solution space, ensuring adherence to the length constraint while achieving sufficiently high generation quality.

\subsection{Length Constraint Score \texorpdfstring{$f(y)$}{}}
\label{sec:length}
Given a word counter $\operatorname{\mathtt{Len}}(\cdot)$, such as the \exper{Nltk} word tokenization function \cite{bird2009natural}, the deviation of the \exper{Llm} generated contents $y$ from the target length $\ell$ can be quantitatively measured using the Manhattan distance, which is:
\begin{equation}
\mathscr{D}(y,\ell) = \vert \operatorname{\mathtt{Len}}(y) - \ell\vert.
\end{equation}

Considering the target of our sampling process, the combination of $f(y)P(y|x)$, we observe a resemblance to the constrained optimization problem, where $\log f(y)$ can be interpreted as a constraint and $\log P(y|x)$ serves as an objective to be maximized. Our target is similar to a Lagrangian function $f(y)P(y|x)\propto \log P(y|x) + \lambda \log f(y)$ with the Lagrange multiplier $\lambda=1$. Furthermore, our sampling process can be seen as analogous to exterior optimization, where the proposed candidates $y_i$ are always unable to satisfy the constraints during each iteration prior to the termination of the loop. Therefore, the constraint function $f(y)$ needs to incentivize candidates that approximate the length constraint, and terminate the iterations with a significantly large reward when the length constraint is satisfied. Thus our length constraint score is defined as follows. For an exact target length $\ell>0$:
\begin{equation}
    f(y) = \frac{1}{\mathscr{D}(y,\ell)}.
\end{equation}
For an interval of target length $[\ell_1, \ell_2]$:
\begin{equation}
    f(y) = \left\{\begin{aligned}
    \frac{1}{\mathscr{D}(y,\ell_1)},&\qquad  y < \ell_1\\
    +\infty,&\quad \ell_1\leq y\leq\ell_2\\
    \frac{1}{\mathscr{D}(y,\ell_2)},&\qquad y>\ell_2 
    \end{aligned}\right.,
\end{equation}
where $0<\ell_1<\ell_2\leq+\infty$ and $\ell_2$ can be $+\infty$. 

\subsection{Probability Distribution of \exper{Llm} \texorpdfstring{$P(y|x)$}{}}
\label{sec:target}
We can obtain the responses generated by black-box \exper{Llm}s such as \exper{Gpt}-4 \cite{achiam2023gpt}, i.e., sampling from their distributions $y\sim P(y|x)$,
\begin{template}{$y_0\sim P(y|x)$}
\texttt{[USER]:} Answer the following instruction using $\{\ell\}$ words or less: $\{x\}$ \vspace{3pt}\\
\texttt{[ASSISTANT]:} Answer: $\{y_{0}\}$
\end{template}
\noindent which is easily accomplished with a simple instruction.
However, we are unable to access their internal parameters or underlying probability distributions. Consequently, it is intractable to verify the probability density $P(y|x)$ of specific samples $y$. 

To address this issue, we employ the LLM-as-a-Judge approach \cite{chiang2023vicuna, zheng2023judging, dubois2024alpacafarm} as a solution. Leveraging the advanced understanding, reasoning, and mathematical capabilities of the model, we require \exper{Llm}s to score samples generated by themselves, thus implicitly estimating their probability density distributions. Besides, we predefine a series of perspectives to unify the scoring mode for our tasks. For the abstractive summarization task, we measure the
\textit{information coverage}, \textit{linguistic} \textit{fluency}, \textit{conciseness}, \textit{logical coherence}, and \textit{faithfulness} of the generated summaries. For instruction following, we measure the response with \textit{helpfulness}, \textit{relevance}, \textit{accuracy}, \textit{depth}, \textit{creativity}, and \textit{level of detail}. Denoting the score function as $\phi(y|x)$, we get the estimated probability distribution as:
\begin{equation}
    P(y|x) \simeq \frac{\phi(y|x)}{\int\phi(y|x)\mathrm{d}y}.
\end{equation}
Similarly to the \cref{eq:A_constraint}, we can cancel the calculation of the normalization constant $\int \phi(y|x)\mathrm{d}y$. Since the \cref{eq:A_constraint} requires the calculation of the ratio $P(y_i|x)\div P(y_{i \minus 1}|x)$, where $y_i$ and $y_{i \minus 1}$ are both sampled from the target \exper{Llm}, we can further refine the scoring function by employing a pairwise comparison function $\Phi(y_i,y_{i \minus 1}|x)$ that
\begin{equation}
    \label{eq:target_distribution}
    \frac{P(y_i|x)}{P(y_{i \minus 1}|x)} \simeq \Phi(y_i,y_{i \minus 1}|x)\simeq\frac{\phi(y_i|x)}{\phi(y_{i \minus 1}|x)},
\end{equation}
where the pairwise score can discern subtle differences between the sample pair $(y_i,y_{i \minus 1})$ at adjacent iteration steps. In addition, the pairwise function will produce scores with less fluctuation than the absolute one $\phi(y|x)$ \cite{zheng2023judging}.

\subsection{Proposal Distribution \texorpdfstring{$p(y_i|y_{i \minus 1},x)$}{}}
\label{sec:proposal}

The proposal distribution $p(y_i|y_{i \minus 1},x)$ plays a pivotal role, as it directly influences the efficiency and quality of the sampling process.
Since the probability distribution of \exper{Llm}, $P(y|x)$, is estimated by itself (\S \ref{sec:target}), requiring \exper{Llm}s to further approximate the probability density of the proposal distribution $p(y_{i}|y_{i \minus 1},x)$ not only introduces additional complexity, but also amplifies the estimation errors.
Therefore, we impose a symmetry constraint \cite{chib1995understanding, haario2001adaptive} on the design of the proposal distribution, which is $q(y_{i}|y_{i \minus 1},x) = q(y_{i \minus 1}|y_{i},x)$. And the acceptance distribution (\cref{eq:A}) reduces to
\begin{equation}
    \label{eq:A_simp}
    \mathcal{A}(y_{i \minus 1}\!\rightarrow\!y_i) =\min\left(1, \frac{f(y_i)P(y_i|x)}{f(y_{i \minus 1})P(y_{i \minus 1}|x)}\right).
\end{equation}
Therefore, we specifically design time-unbiased instructions to make \exper{Llm} satisfy the symmetry constraints as much as possible. The detailed prompt template for \exper{Llm}s is as follows:
\begin{template}{$y_i\sim p(y_i|y_{i \minus 1},x)$}
\texttt{[USER]:} Answer the following instruction using $\{\ell\}$ words or less: $\{x\}$ \vspace{3pt}\\
\texttt{[ASSISTANT]:} Answer: $\{y_{i \minus 1}\}$
\tcblower
\texttt{[USER]:} Please generate a new answer based on the previous one: \vspace{3pt}\\
\texttt{[ASSISTANT]:} Answer: $\{y_{i}\}$
\end{template}
\noindent where $y_i$ and $y_{i \minus 1}$ are equivalent and interchangeable in the semantics of this template.

However, this preliminary Metropolis-Hastings framework, constructed with the current proposal function $p(y_i|y_{i \minus 1},x)$, is not efficient due to following reasons. (1) Intuitively, when generating new candidates, the length signal remains the initial one (\textit{"using $\ell$ words or less"}). 
Without introducing updated length signals, \exper{Llm} may remain trapped in its own errors, unable to converge to improved solutions. (2) From a theoretical perspective, the sampling efficiency and quality will be maximized when the proposal function $p(y_i|y_{i \minus 1},x)$ aligns closely with the target distribution $\pi(y|x)$ \cite{gelman1997weak}. This means that the sampling efficiency decreases as this discrepancy increases.

Therefore, we apply the importance sampling strategy \cite{kahn1953methods, owen2000safe} to improve the proposal distribution.  We define an importance distribution $q(y_i|y_{i \minus 1},x)$ that complies with length constraints, serving as a replacement for the proposal distribution to facilitate accelerated sampling. \Cref{eq:A,eq:A_simp} can be further derived when $y_i\sim q(y_i|y_{i \minus 1},x)$:
\begin{equation}
    \label{eq:upperbound}
    \begin{aligned}
    &\mathcal{A}(y_{i \minus 1}\!\rightarrow\!y_i)\!=\!\frac{p(y_i|y_{i \minus 1},x)}{q(y_i|y_{i \minus 1},x)}\min\left(1, \frac{\pi(y_i|x)}{\pi(y_{i \minus 1}|x)}\right)\\
    &\quad \qquad \leq \min\left(1, \frac{f(y_i)P(y_i|x)}{f(y_{i \minus 1})P(y_{i \minus 1}|x)}\right),
    \end{aligned}
\end{equation}
where $\frac{p(y_i|y_{i \minus 1},x)}{q(y_i|y_{i \minus 1},x)}\!\leq\!1$ and \cref{eq:A_simp} becomes an upper bound of $\mathcal{A}(y_{i \minus 1}\!\rightarrow\!y_i)$. By simply replacing \cref{{alg:propose}} in \cref{alg:framework} with $y_i \sim q(y_i|y_{i \minus 1}|x)$ and calculating the acceptance rate with the upper bound \cref{eq:upperbound}, we can significantly accelerate the sampling process. Although calculating this upper bound may lead to higher acceptance rates, potentially compromising generation quality, the remarkable capabilities of \exper{Llm}s fortunately mitigate this risk to an almost negligible level. In addition, the detailed template for the importance distribution is:
\begin{template}{$y_i\sim q(y_i|y_{i \minus 1},x)$}
\texttt{[USER]:} Answer the following instruction using $\{\ell\}$ words or less: $\{x\}$ \vspace{3pt}\\
\texttt{[ASSISTANT]:} Answer: $\{y_{i \minus 1}\}$
\tcblower
\texttt{[USER]:} The generated answer is too (long / short) at $\{\mathtt{Len}(y)\}$ words.\\
Please (delete / add) $\{\mathscr{D}(y,\ell)\}$ words appropriately based on the previous response:\vspace{3pt}\\
\texttt{[ASSISTANT]:} Answer: $\{y_{i}\}$
\end{template}

\begin{table*}[t]
\centering
\setlength{\tabcolsep}{4mm}{
\resizebox{0.9\textwidth}{!}{
  \begin{tabular}{l:r||c|c|c||c|c|c}
    \hline
    \multicolumn{1}{c}{\textbf{Models}} & \textbf{Samplers} & \textbf{\exper{Acc}$\uparrow$} & \textbf{\exper{L1}$\downarrow$} & \textbf{\exper{L2}$\downarrow$} & \textbf{\exper{Rouge-1}} & \textbf{\exper{Rouge-2}} & \textbf{\exper{Rouge-L}} \\
    \hline
    
    \hline
    \hline
    \multirow{2}{*}{\exper{Llama2}} & \exper{Inst} & $\phantom{00}4.1\%$ & $11.42$ & $15.20$ & $0.37$ & $0.13$ & $0.24$\\
    & \exper{Ours} & $\phantom{0}81.6\%$ & $\phantom{0}0.24$ & $\phantom{0}0.64$ & $0.36$ & $0.12$ & $0.24$\\
    \cline{2-8}
    \multirow{2}{*}{\exper{Qwen2.5}} & \exper{Inst} & $\phantom{00}3.2\%$ & $\phantom{0}8.64$ & $10.83$ & $0.33$ & $0.10$ & $0.21$\\
    & \exper{Ours} & $\phantom{0}86.4\%$ & $\phantom{0}0.18$ & $\phantom{0}0.72$ & $0.33$ & $0.10$ & $0.21$\\
    \cline{2-8}
    \multirow{2}{*}{\exper{Llama3}} & \exper{Inst} & $\phantom{00}9.1\%$ & $\phantom{0}4.78$ & $\phantom{0}6.10$ & $0.39$ & $0.14$ & $0.25$\\
    & \exper{Ours} &$\phantom{0}78.6\%$ & $\phantom{0}0.29$ & $\phantom{0}0.66$ & $0.38$ & $0.14$ & $0.25$\\
    \cline{2-8}
    \multirow{2}{*}{\exper{Llama3.1}} & \exper{Inst} & $\phantom{00}7.7\%$ & $\phantom{0}3.88$ & $\phantom{0}5.10$ & $0.38$ & $0.13$ & $0.24$\\
    & \exper{Ours} & $100.0\%$ & $\phantom{0}0.00$ & $\phantom{0}0.00$ & $0.38$ & $0.13$ & $0.24$\\
    \cline{2-8}
    \multirow{2}{*}{\exper{Gpt}-3.5} & \exper{Inst} & $\phantom{00}5.1\%$ & $\phantom{0}8.29$ & $13.69$ & $0.36$ & $0.12$ & $0.23$\\
    & \exper{Ours} & $\phantom{0}95.0\%$ & $\phantom{0}0.14$ & $\phantom{0}1.11$ & $0.36$ & $0.12$ & $0.23$\\
    \cline{2-8}
    \multirow{2}{*}{\exper{Gpt}-4} & \exper{Inst} & $\phantom{0}15.7\%$ & $\phantom{0}2.10$ & $\phantom{0}2.67$ & $0.36$ & $0.12$ & $0.23$\\
    & \exper{Ours} & $\phantom{0}99.2\%$ & $\phantom{0}0.01$ & $\phantom{0}0.12$ & $0.36$ & $0.12$ & $0.23$\\
    \hline
    \hline

    \hline
  \end{tabular}
  }}
  \caption{Results of the length control on the CNN/DailyMail dataset. \exper{Inst} is the baseline response with length-guided instructions. \exper{Ours} represents our iterative sampling framework.}
  \label{tab:cnndm}
\end{table*}

It should be noted that our method can perform parallel sampling as long as the corresponding \exper{Llm} supports it, further improving the control efficiency.

\section{Experiments}
\subsection{Experimental Setup}
\paragraph{Datasets} For exact length control, we utilize the CNN/DailyMail dataset (\exper{CnnDm}, \citet{nallapati2016abstractive}), where the length instruction $\ell$ is extracted from the references. For length-interval control, we use the Alpaca-Eval-LI (\exper{Alpaca}) and MT-Bench-LI (\exper{MtBench}) datasets \cite{yuan2024following}, which are derived from the Alpaca-Eval dataset \cite{dubois2024alpacafarm} and the MT-Bench dataset \cite{zheng2023judging}. The length interval instructions are already provided in the dataset, where $\ell_1=0$ and $\ell_2$ is the length of the reference response.

\paragraph{\exper{Llm}s} We evaluate the effectiveness of our framework in the latest \exper{Llm}s, including Llama-2-7B (\exper{Llama2}, \citet{touvron2023llama2}, Qwen-2.5-7B (\exper{Qwen2.5}, \citet{qwen2.5}), Llama-3/3.1-8B (\exper{Llama3/3.1}, \citet{dubey2024llama}), and \exper{Gpt-3.5/4} \cite{achiam2023gpt}. For white-box \exper{Llm}s like \exper{Llama}, we use them as black-box models, where the maximum iteration trial is $5$ with a beam size of $16$. For black-box models based on APIs like \exper{Gpt-4}, we set the maximum iteration trial as $15$ without parallel sampling.

\begin{table}[t]
  \vspace{5pt}
  \setlength{\tabcolsep}{2.8pt}
  \centering
  \begin{tabular}{l:r||c|c|c||c}
    \hline
    \multicolumn{1}{c}{\textbf{Models}} & \textbf{Samp.} & \textbf{\exper{Acc}$\uparrow$} & \textbf{\exper{L1}$\downarrow$} & \textbf{\exper{L2}$\downarrow$} & \textbf{\exper{Win.}}$\uparrow$ \\
    \hline
    
    \hline
    \hline
    \multicolumn{6}{l}{\textbf{\exper{Alpaca-Eval-LI}}}\\
    \hline
    \multirow{2}{*}{\exper{\footnotesize{Llama3}}} & \exper{Inst} & $\phantom{0}92.2\%$ & $\phantom{0}1.48$ & $\phantom{0}9.12$ & $76.5\%$ \\
    & \exper{Ours} & $\phantom{0}99.8\%$ & $\phantom{0}0.02$ & $\phantom{0}0.05$ & $83.5\%$\\
    \cline{2-6}
    \multirow{2}{*}{\exper{\scriptsize{Llama3.1}}} & \exper{Inst} & $\phantom{0}91.6\%$ & $\phantom{0}2.47$ & $15.64$ & $71.6\%$ \\
    & \exper{Ours} & $\phantom{0}99.8\%$ & $\phantom{0}0.06$ & $\phantom{0}1.69$ & $76.7\%$\\
    \cline{2-6}
    \multirow{2}{*}{\footnotesize{\exper{Gpt}-3.5}} & \exper{Inst} & $\phantom{0}91.5\%$ & $\phantom{0}1.16$ & $\phantom{0}4.92$ & $57.0\%$ \\
    & \exper{Ours} & $100.0\%$ & $\phantom{0}0.00$ & $\phantom{0}0.00$ & $65.3\%$\\
    \cline{2-6}
    \multirow{2}{*}{\footnotesize{\exper{Gpt}-4}} & \exper{Inst} & $\phantom{0}37.2\%$ & $21.38$ & $37.61$ & $30.2\%$ \\
    & \exper{Ours} & $\phantom{0}99.2\%$ & $\phantom{0}0.02$ & $\phantom{0}0.17$ & $92.0\%$\\
    \hline
    \hline

    \hline
    \multicolumn{6}{l}{\textbf{\exper{MT-Bench-LI}}}\\
    \hline
    \multirow{2}{*}{\exper{\footnotesize{Llama3}}} & \exper{Inst} & $\phantom{0}78.8\%$ & $\phantom{0}2.80$ & $\phantom{0}7.73$ & $41.1\%$ \\
    & \exper{Ours} & $100.0\%$ & $\phantom{0}0.00$ & $\phantom{0}0.00$ & $42.1\%$\\
    \cline{2-6}
    \multirow{2}{*}{\exper{\scriptsize{Llama3.1}}} & \exper{Inst} & $\phantom{0}80.4\%$ & $10.15$ & $60.71$ & $35.2\%$ \\
    & \exper{Ours} & $\phantom{0}98.8\%$ & $\phantom{0}0.73$ & $\phantom{0}7.12$ & $42.9\%$\\
    \cline{2-6}
    \multirow{2}{*}{\footnotesize{\exper{Gpt}-3.5}} & \exper{Inst} & $\phantom{0}87.9\%$ & $\phantom{0}2.51$ & $\phantom{0}9.46$ & $24.6\%$ \\
    & \exper{Ours} & $\phantom{0}98.6\%$ & $\phantom{0}0.09$ & $\phantom{0}0.73$ & $27.3\%$\\
    \cline{2-6}
    \multirow{2}{*}{\footnotesize{\exper{Gpt}-4}} & \exper{Inst} & $\phantom{0}54.7\%$ & $13.99$ & $29.16$ & $27.4\%$ \\
    & \exper{Ours} & $\phantom{0}98.8\%$ & $\phantom{0}0.05$ & $\phantom{0}0.41$ & $63.7\%$\\
    \hline
    \hline

    \hline
  \end{tabular}
  \caption{Results of the length control on the Alpaca-Eval-LI dataset and the MT-Bench-LI dataset.}
  \label{tab:alpaca}
\end{table}

\paragraph{Evaluation Metrics}
We use several metrics to estimate the effect of the length control. \textbf{\exper{Acc}}uracy measures the ratio of generated contents that are fully in accordance with the length constraint. Given $N$ generated contents, \textbf{\exper{L1}} measures the average Manhattan distance $\frac{1}{N}\sum_y\vert \operatorname{\mathtt{Len}}(y) - \ell\vert$ and \textbf{\exper{L2}} measures the average Euclidean distance $\sqrt{\frac{1}{N}\sum_y\vert \operatorname{\mathtt{Len}}(y) - \ell\vert^2}$. For quality evaluation of the summary task, we use the classic score \textbf{\exper{Rouge}} \cite{lin-2004-rouge}. For the instruction following tasks, we use the length-instructed \textbf{\exper{Win}}rate \cite{yuan2024following}, where responses are compared pairwise with baselines. The winner is determined by both the quality of the responses provided by LLM-as-a-Judge \cite{zheng2023judging}, and the adherence to the length constraints. If the response exceeds the length constraint, it is automatically lost.

\begin{table}[t]
  \setlength{\tabcolsep}{3pt}
  \centering
  \begin{tabular}{c||c|c|c||c}
    \hline
    \textbf{Trials} &\textbf{\exper{Acc}$\uparrow$} & \textbf{\exper{L1}$\downarrow$} & \textbf{\exper{L2}$\downarrow$} & \textbf{\exper{Rouge-(1/2/L)}}\\
    \hline
    
    \hline
    \hline
    \multicolumn{5}{l}{\textbf{\exper{Qwen2.5}}}\\
    \hline
    $0$ & $\phantom{00}3.2\%$ & $8.64$ & $10.83$ & $0.33/0.10/0.21$\\
    $1$ & $\phantom{0}25.4\%$ & $3.58$ & $\phantom{0}5.80$ & $0.33/0.10/0.21$\\
    $2$ & $\phantom{0}52.8\%$ & $0.96$ & $\phantom{0}2.07$ & $0.33/0.10/0.21$\\
    $3$ & $\phantom{0}70.6\%$ & $0.51$ & $\phantom{0}1.55$ & $0.33/0.10/0.21$\\
    $4$ & $\phantom{0}79.1\%$ & $0.29$ & $\phantom{0}1.16$ & $0.33/0.10/0.21$\\
    $5$ & $\phantom{0}86.4\%$ & $0.18$ & $\phantom{0}0.72$ & $0.33/0.10/0.21$\\
    \hline
    \hline

    \hline
    \multicolumn{5}{l}{\textbf{\exper{Llama3.1}}}\\
    \hline
    $0$ & $\phantom{00}7.7\%$ & $3.88$ & $\phantom{0}5.10$ & $0.38/0.13/0.24$\\
    $1$ & $\phantom{0}86.4\%$ & $0.18$ & $\phantom{0}0.55$ & $0.38/0.13/0.24$\\
    $2$ & $\phantom{0}99.2\%$ & $0.04$ & $\phantom{0}0.28$ & $0.38/0.13/0.24$\\
    $3$ & $\phantom{0}99.8\%$ & $0.01$ & $\phantom{0}0.03$ & $0.38/0.13/0.24$\\
    $4$ & $100.0\%$ & $0.00$ & $\phantom{0}0.00$ & $0.38/0.13/0.24$\\
    $5$ & $100.0\%$ & $0.00$ & $\phantom{0}0.00$ & $0.38/0.13/0.24$\\
    \hline
    \hline

    \hline
  \end{tabular}
  \caption{Analysis of the iteration trial on the \exper{CnnDm} dataset, where the beam size is $16$.}
  \label{tab:trial}
\end{table}

\begin{table}[t]
  \setlength{\tabcolsep}{3pt}
  \centering
  \begin{tabular}{c||c|c|c||c}
    \hline
    \textbf{Beams} & \textbf{\exper{Acc}$\uparrow$} & \textbf{\exper{L1}$\downarrow$} & \textbf{\exper{L2}$\downarrow$} & \textbf{\exper{Rouge-(1/2/L)}} \\
    \hline
    
    \hline
    \hline
    \multicolumn{5}{l}{\textbf{\exper{Qwen2.5}}}\\
    \hline
    $\phantom{0}0$ & $\phantom{00}3.2\%$ & $8.64$ & $10.83$ & $0.33/0.10/0.21$\\
    $\phantom{0}1$ & $\phantom{0}24.6\%$ & $2.41$ & $\phantom{0}4.02$ & $0.33/0.10/0.21$\\
    $\phantom{0}2$ & $\phantom{0}38.7\%$ & $1.75$ & $\phantom{0}3.69$ & $0.33/0.10/0.21$\\
    $\phantom{0}4$ & $\phantom{0}57.1\%$ & $0.82$ & $\phantom{0}1.93$ & $0.32/0.10/0.21$\\
    $\phantom{0}8$ & $\phantom{0}72.6\%$ & $0.46$ & $\phantom{0}1.49$ & $0.32/0.10/0.20$\\
    $16$ & $\phantom{0}86.4\%$ & $0.18$ & $\phantom{0}0.72$ & $0.33/0.10/0.21$\\
    \hline
    \hline

    \hline
    \multicolumn{5}{l}{\textbf{\exper{Llama3.1}}}\\
    \hline
    $\phantom{0}0$ & $\phantom{00}7.7\%$ & $3.88$ & $\phantom{0}5.10$ & $0.38/0.13/0.24$\\
    $\phantom{0}1$ & $\phantom{0}93.3\%$ & $0.14$ & $\phantom{0}0.88$ & $0.37/0.13/0.24$\\
    $\phantom{0}2$ & $\phantom{0}98.9\%$ & $0.02$ & $\phantom{0}0.12$ & $0.37/0.13/0.24$\\
    $\phantom{0}4$ & $\phantom{0}99.7\%$ & $0.01$ & $\phantom{0}0.05$ & $0.38/0.14/0.25$\\
    $\phantom{0}8$ & $100.0\%$ & $0.00$ & $\phantom{0}0.00$ & $0.38/0.13/0.24$\\
    $16$ & $100.0\%$ & $0.00$ & $\phantom{0}0.00$ & $0.38/0.13/0.24$\\
    \hline
    \hline

    \hline
  \end{tabular}
  \caption{Analysis of the beam size on the \exper{CnnDm} dataset, where the iteration trial is $5$.}
  \label{tab:beam}
\end{table}

\subsection{Main Results}
The detailed comparisons between the baselines and our framework are demonstrated in \Cref{tab:cnndm,tab:alpaca}. \Cref{tab:cnndm} presents the results of length control experiments conducted on the CNN/DailyMail dataset. Our method (denoted as \exper{Ours}) demonstrates significant improvements over the baseline instruction method (\exper{Inst}) across all evaluated models and length-related metrics.
Specifically, we achieve near-perfect or perfect accuracy (\exper{Acc}), with values exceeding $95\%$ for the most advanced \exper{Llm}s (\exper{Llama3.1}, \exper{Gpt-3.5}, and \exper{Gpt-4}), while the baselines struggle with accuracy values below $16\%$. Furthermore, our approach exhibits substantially lower errors of \exper{L1} and \exper{L2}, indicating precise adherence to the target lengths. For example, on \exper{Llama3.1}, our framework achieves an accuracy of $100\%$, demonstrating flawless length control. Similarly, we attain a $99.2\%$ accuracy on \exper{Gpt-4}, reducing the \exper{L1} and \exper{L2} errors to 0.01 and 0.12, respectively. Beyond the significant improvement in length control, our method introduces almost no degradation in generation quality, where the \exper{Rouge} metrics remain almost the same to the instruction-based baselines.

\Cref{tab:alpaca} evaluates the performance of our method on Alpaca-Eval-LI and MT-Bench-LI datasets. Although these two datasets are relatively easier compared to the exact length control task, the performance improvement (with an accuracy increase of at least $7.6\%$) brought about by our method compared to the baseline is significant, confirming the consistent superiority of our framework across different benchmarks. In addition, the \exper{Llm} judged pairwise \exper{Win}rate of our approach improves.
These results highlight the effectiveness of our iterative sampling framework in achieving robust and accurate length control across diverse \exper{Llm}s.

\begin{table}[t]
  \vspace{10pt}
  \setlength{\tabcolsep}{3pt}
  \centering
  \begin{tabular}{l||c|c|c||c}
    \hline
    \textbf{Samp.} & \textbf{\exper{Acc}$\uparrow$} & \textbf{\exper{L1}$\downarrow$} & \textbf{\exper{L2}$\downarrow$} & \textbf{\exper{Rouge-(1/2/L)}} \\
    \hline
    
    \hline
    \hline
    \exper{Inst} & $\phantom{0}7.7\%$ & $3.88$ & $5.10$ & $0.38/0.13/0.24$\\
    \exper{Rand} & $38.8\%$ & $1.18$ & $1.85$ & $0.38/0.14/0.24$\\
    \exper{Mh} & $40.2\%$ & $1.47$ & $3.20$ & $0.36/0.13/0.23$\\
    \exper{Mh+Is} & $93.3\%$ & $0.14$ & $0.88$ & $0.37/0.13/0.24$\\
    \hline
    \hline

    \hline
  \end{tabular}
  \caption{Ablation study of \exper{Llama3.1} on \exper{CnnDm}, where the iteration trial is $5$ and the beam size is $1$.}
  \label{tab:ablation}
\end{table}

\subsection{Analyses}
\label{sec:analyze}
We analyze the hyperparameters of our framework, the number of iteration trials and the beam size, which are illustrated in \Cref{tab:trial,tab:beam}. Both hyperparameters are used to expand and explore the sampling space, with larger iteration trials demanding greater time overhead and larger beam sizes incurring higher space costs.
As the sampling space reduces, we can observe that the influence of length control progressively decreases. In particular, this reduction is non-linear, with the rate of decline accelerating significantly.
Considering both \exper{Llama3.1} and \exper{Qwen2.5}, the marginal effect of expanding the sampling space decreases more rapidly for the stronger model.
In addition, comparing the two tables, we can observe that the iteration trial number contributes more to the control effect than the beam size.
With a smaller sampling space of $2$ beams $\times$ $5$ trials, the accuracy ($38.7\%$ of \exper{Qwen2.5} and $98.9\%$ of \exper{Llama3.1}) outperforms the situation with $16$ beams $\times$ $1$ trial ($25.4\%$ of \exper{Qwen2.5} and $86.4\%$ of \exper{Llama3.1}).

\subsection{Ablation Study}
\label{sec:ablation}
\Cref{tab:ablation} presents an ablation study evaluating the performance of different sampling strategies for \exper{Llama3.1} on the CNN/DailyMail dataset. We examine four sampling strategies: (1) \exper{Inst} is the instruction following baseline without iterations; (2) \exper{Rand} extends the baseline to resample at each iteration and retains the best one; (3) \exper{Mh} is our initial version of the Metropolis-Hastings framework that resamples with the proposal distribution $y\sim p(y_i|y_{i-1},x)$ during each iteration; and (4) \exper{Mh+Is} is our complete method which replaces the proposal distribution with the importance distribution $q(y_i|y_{i-1},x)$. We set the beam size to $1$, because sampling a batch of initial states $y_0\!\sim\! P(y|x)$ is actually the \exper{Rand} method and we want to eliminate this influence. Experimental results show that with the powerful instruction following capabilities of \exper{Llm}s, random sampling of more candidates can achieve good control effects. However, the original Metropolis-Hastings method does not make the process more efficient and is sometimes even worse than random sampling. By replacing the proposal distribution with an importance sampling strategy, our method achieves significant improvements.

\begin{table}[t]
  \setlength{\tabcolsep}{3pt}
  \centering
  \begin{tabular}{l:c|c||c|c|c}
    \hline
    \multicolumn{1}{c}{\textbf{Task}} & \multicolumn{1}{c|}{\textbf{Samp.}} & \textbf{\exper{\small{Steps}}$\downarrow$} & \textbf{\exper{Acc}$\uparrow$} &\textbf{\exper{L1}$\downarrow$} & \textbf{\exper{L2}$\downarrow$} \\
    \hline
    
    \hline
    \hline
    \multirow{3}{*}{\small{\exper{CnnDm}}} & \exper{Rand} & $18.6$ & $\phantom{0}95.3\%$ & $0.06$ & $0.35$\\
    & \exper{Mh} & $17.1$ & $\phantom{0}96.0\%$ & $0.34$ & $1.85$\\
    & \exper{Mh+Is} & $\phantom{0}2.4$ & $100.0\%$ & $0.00$ & $0.00$\\
    \cline{2-6}
    \multirow{3}{*}{\small{\exper{Alpaca}}} & \exper{Rand} & $\phantom{0}0.6$ & $\phantom{0}98.0\%$ & $0.21$ & $1.71$\\
    & \exper{Mh} & $\phantom{0}0.9$ & $\phantom{0}95.3\%$ & $0.32$ & $2.67$\\
    & \exper{Mh+Is} & $\phantom{0}0.1$ & $100.0\%$ & $0.00$ & $0.00$\\
    \cline{2-6}
    \multirow{3}{*}{\small{\exper{MtBench}}} & \exper{Rand} & $\phantom{0}3.0$ & $\phantom{0}97.9\%$ & $0.33$ & $3.26$\\
    & \exper{Mh} & $\phantom{0}3.3$ & $\phantom{0}96.7\%$ & $0.42$ & $4.88$\\
    & \exper{Mh+Is} & $\phantom{0}0.8$ & $\phantom{0}99.5\%$ & $0.06$ & $0.85$\\

    \hline
    \hline

    \hline
  \end{tabular}
  \caption{Convergence steps on \exper{Llama3.1}.}
  \label{tab:convergence_llama}
\end{table}

\begin{table}[t]
  \setlength{\tabcolsep}{3pt}
  \centering
  \begin{tabular}{l:c|c||c|c|c}
    \hline
    \multicolumn{1}{c}{\textbf{Task}} & \multicolumn{1}{c|}{\textbf{Samp.}} & \textbf{\exper{\small{Steps}}$\downarrow$} & \textbf{\exper{Acc}$\uparrow$} &\textbf{\exper{L1}$\downarrow$} & \textbf{\exper{L2}$\downarrow$} \\
    \hline
    
    \hline
    \hline
    \multirow{3}{*}{\small{\exper{CnnDm}}} & \exper{Rand} & $2.6$ & $93.8\%$ & $0.06$ & $\phantom{0}0.29$\\
    & \exper{Mh} & $2.5$ & $91.4\%$ & $0.09$ & $\phantom{0}0.58$\\
    & \exper{Mh+Is} & $1.0$ & $98.0\%$ & $0.02$ & $\phantom{0}0.14$\\
    \cline{2-6}
    \multirow{3}{*}{\small{\exper{Alpaca}}} & \exper{Rand} & $2.5$ & $77.3\%$ & $2.52$ & $\phantom{0}6.54$\\
    & \exper{Mh} & $3.0$ & $93.6\%$ & $3.02$ & $\phantom{0}8.67$\\
    & \exper{Mh+Is} & $0.4$ & $98.2\%$ & $0.09$ & $\phantom{0}0.81$\\
    \cline{2-6}
    \multirow{3}{*}{\small{\exper{MtBench}}} & \exper{Rand} & $1.3$ & $83.7\%$ & $2.57$ & $\phantom{0}7.89$\\
    & \exper{Mh} & $1.8$ & $78.3\%$ & $4.09$ & $14.71$\\
    & \exper{Mh+Is} & $0.1$ & $99.8\%$ & $0.01$ & $\phantom{0}0.09$\\

    \hline
    \hline

    \hline
  \end{tabular}
  \caption{Convergence steps on \exper{Gpt-4}.}
  \label{tab:convergence_gpt}
\end{table}

\subsection{Convergence Study}
Furthermore, we analyze the accurate convergence speed of different sampling methods in \Cref{tab:convergence_llama,tab:convergence_gpt}. We set the beam size to $1$ as in \cref{sec:ablation} and the maximum iteration step for each case is $100$ for \exper{Llama3.1} and $15$ for \exper{Gpt-4}. We report the average iteration \textbf{\exper{Steps}} required to satisfy the length constraints, which excludes the first sampling step $y_0\sim P(y|x)$. We observe that different models have different convergence steps for different tasks. In general, precise length control tasks are more difficult and require more iterations. Even so, we achieve an almost perfect control effect with only $2.4$ iteration steps on average for \exper{Llama3.1}. We even only need an average of $0.1$ iterations for \exper{Llama3.1} to perform perfect control on the Alpaca-Eval-LI dataset. For \exper{Gpt-4}, we only need $1.0$ iterations at most on average to obtain good control results. Therefore, our framework can achieve extremely effective length control performance with acceptable time overhead.

\section{Conclusion}
We propose a novel length controllable sampling framework for black-box models and verify the effectiveness with experiments and analyses. Our study confirms that an almost perfect length control can be achieved on \exper{Llm}s, which is of great significance to improve their instruction following ability. In addition, although our framework performs well, its sampling efficiency and generation effect are affected by the capabilities of \exper{Llm} itself. Fortunately, with the rapid development of \exper{Llm}s, this concern will gradually disappear. 
Its worth noting that we do not directly compare with the length training methods, because (1) the black-box models are not trainable, and (2) the training methods are based on specific datasets and possess some data bias, which is contrary to the objective of a more generalized length control. We hope to explore more efficient and general length control schemes in our future studies.

\section*{Limitations}
Despite the promising results demonstrated in our experiments, our method has some limitations that merit further discussion:
\begin{itemize}[leftmargin=0pt]
    \item \textbf{Inference Overhead:}\\
    Our approach introduces additional inference overhead due to the iterative nature of the method. Although the experimental results show that satisfactory results can often be achieved in $2$ iterations for advanced models such as \exper{Llama3.1}. However, more iteration steps are required for more difficult scenarios or weaker \exper{Llm}s. This additional computational cost may present challenges for large-scale batch generation tasks where inference speed is critical. Future research could explore optimization techniques to reduce the number of iterations required or design lightweight variants to better suit the high-throughput applications.
    \item \textbf{Dependency on Instruction Following Abilities:}\\
    The performance of our method is highly dependent on the instruction following capabilities of the underlying model. For state-of-the-art \exper{Llm}s such as \exper{Llama3.1} and \exper{Gpt-4}, fewer iterations are typically sufficient to achieve satisfactory results. However, when applied to models with less robust instruction-following abilities, the number of iterations required may increase significantly, potentially affecting efficiency. Addressing this limitation could involve developing methods to enhance instruction alignment for less capable models or incorporating external mechanisms to mitigate the dependency on instruction following abilities.
\end{itemize}

\noindent Considering our experiments, the limitations are:
\begin{itemize}[leftmargin=0pt]
\item \textbf{Baselines:} We do not directly compare with training methods for length control because: (1) our framework is dedicated to black-box \exper{Llm}s, which is not trainable; (2) length instructions have already been incorporated in the supervised fine-tuning stage of \exper{Llm}s, which means \exper{Llm}s themselves are length trainable baselines; (3) the training
methods are based on specific datasets and possess some data bias, which is contrary to the objective of a more generalized length control; and (4) large-scale training of length instructions on \exper{Llm}s such as \exper{Llama3.1} requires a lot of computing resources that we cannot currently afford.
\item \textbf{Models:} Currently, we only test the most widely used \exper{Llm}s. Due to the limitations of computing resources and costs, we are unable to test white-box models with larger parameters (such as $70$B), nor can we afford the test of other API-based black-box \exper{Llm}s on a large scale.
\end{itemize}

\section*{Ethics Statement}
This research focuses on controlling the output length of \exper{Llm}s to address practical usability and fairness concerns in various applications, such as summarization, dialogue systems, and content generation. By enabling precise length control, this work aims to enhance user experience, ensure relevance, and reduce unintended biases introduced by excessively verbose or overly concise outputs.

We recognize the potential ethical risks associated with the misuse of controlled generation, such as the creation of misleading or harmful content tailored to specific lengths. To mitigate such risks, our methodology emphasizes transparency, reproducibility, and alignment with ethical guidelines in AI development. Additionally, we advocate for integrating robust content moderation mechanisms in downstream applications to safeguard against unintended consequences.

This research was conducted following established ethical standards, ensuring that the datasets used respect privacy and are free of harmful biases to the best of our ability. Future work will further explore the societal implications of this technology, ensuring its responsible and equitable deployment.
\bibliography{custom}

\newpage
\appendix
\section{Prompt Templates}
\label{app:template}
\subsection{Initial States}
For the abstractive summarization task with exact length constraints, we randomly choose an example $(x_c,y_c,\ell_c)$ from the training set as an one-shot demonstration for \exper{Llm}s, because the chat \exper{Llm}s are not specifically trained for the output mode of summary tasks. The detailed template is:
\begin{template}{$y_0\sim P(y|x)$}
\texttt{[SYSTEM]:} You are a powerful abstractive summarizer. \vspace{3pt}\\
\texttt{[USER]:} Document: $\{x_c\}$\\ Based on the previous document, provide a high-quality summary in exactly $\{\ell_c\}$ words:\vspace{3pt}\\
\texttt{[ASSISTANT]:} Summary: $\{y_{c}\}$
\tcblower
\texttt{[USER]:} Document: $\{x\}$\\
Based on the previous document, provide a high-quality summary in exactly $\{\ell\}$ words:\vspace{3pt}\\
\texttt{[ASSISTANT]:} Summary: $\{y_{0}\}$
\end{template}
\noindent For instruction following tasks with length intervals, we directly use zero-shot with the template: 
\begin{template}{$y_0\sim P(y|x)$}
\texttt{[USER]:} Answer the following instruction using $\{\ell\}$ words or less: $\{x\}$ \vspace{3pt}\\
\texttt{[ASSISTANT]:} Answer: $\{y_{0}\}$
\end{template}

\subsection{Probability Densities of Current States}
We demonstrate the detailed $\{Criteria\}$ of evaluation for different tasks. For abstractive summarization: we score the generated summaries in $5$ dimensions on a scale of 1-10.
{\color{gray}
\begin{enumerate}[leftmargin=13pt]
    \item \textbf{Information Coverage:} Does the summary include the most important and critical information from the document?
    \item \textbf{Linguistic Fluency:} Are the sentences in the summary fluent, natural, and grammatically correct?
    \item \textbf{Conciseness:} Does the summary avoid redundancy while retaining key information?
    \item \textbf{Logical Coherence:} Is the summary well-structured with clear and logical flow?
    \item \textbf{Faithfulness:} Does the summary accurately reflect the facts in the original document without adding false or misleading information?
\end{enumerate}
}

The evaluation $\{Criteria\}$ for the general instruction following task is of $6$ dimensions:
{\color{gray}
\begin{enumerate}[leftmargin=13pt]
    \item \textbf{Helpfulness:} Does the response directly address the instruction and provide meaningful assistance?
    \item \textbf{Relevance:} Does the response stay on topic and avoid unnecessary or unrelated information?
    \item \textbf{Accuracy:} Is the information in the response factually correct and free of errors?
    \item \textbf{Depth:} Does the response demonstrate a deep understanding of the topic, including nuanced explanations where relevant?
    \item \textbf{Creativity:} Does the response display originality, creativity, or a unique approach to addressing the instruction?
    \item \textbf{Level of Detail:} Is the response sufficiently detailed, providing comprehensive and thorough explanations where necessary?
\end{enumerate}
}
Following the setting of MT-Bench, we set a special evaluation $\{Criteria\}$ for math-related instruction following tasks such as reasoning, math and coding, which is described below. 
{\color{gray}
\begin{enumerate}[leftmargin=13pt]
    \item \textbf{Correctness:} Is the answer logically sound, factually accurate, and free from errors?
    \item \textbf{Helpfulness:} Does the response directly address the instruction and provide meaningful assistance?
    \item \textbf{Clarity:} Is the response well-structured and easy to understand?
    \item \textbf{Efficiency:} Does the response provide an optimal solution without unnecessary complexity?
    \item \textbf{Completeness:} Does the response fully cover the instruction's requirements and edge cases?
    \item \textbf{Robustness:} Can the response handle ambiguity or complexity in the instruction?
\end{enumerate}
}
We formalize the output to facilitate the extraction of key information, where the $\{Format\}$ is
\begin{format}{}
\#\#\#\# Response 1:\\
1. Information Coverage:: [Score]/10\\
2. Linguistic Fluency: [Score]/10\\
$\dots\dots$\\
**Overall Score:** [Total Score]/50\\
\#\#\#\# Response 2:\\
1. Information Coverage:: [Score]/10\\
2. Linguistic Fluency: [Score]/10\\
$\dots\dots$\\
**Overall Score:** [Total Score]/50\\
\#\#\# Conclusion:\\
- **Better Response:** [Response 1/Response 2].\\
- **Score Ratio (Response 1 ÷ Response 2):** [Ratio, rounded to two decimal places].
\end{format}
\noindent We calculate the \cref{eq:target_distribution} via 
\begin{equation}
    \begin{aligned}
        \frac{P(y_i|x)}{P(y_{i \minus 1}|x)} &\simeq \frac{\phi(y_i|x)}{\phi(y_{i \minus 1}|x)}\\
        &= \frac{\text{Score of Response }1}{\text{Score of Response }2}.
    \end{aligned}
\end{equation}
Therefore, the prompt templates for estimating the target probability density are:

\begin{template}{$\Phi(y_i,y_{i-1}|x)$}
\texttt{[SYSTEM]:} You are a powerful evaluator for abstractive summarization. \vspace{3pt}\\
\texttt{[USER]:} I need to compare and evaluate the quality of two summaries generated for a given document. Please provide a quantitative assessment of their performance based on the criteria below.\\
Document: $\{x\}$\\
Summary 1: $\{y_{i}\}$\\
Summary 2: $\{y_{i-1}\}$\\
Evaluation Criteria (each scored on a scale of 1-10, with 10 being the best): $\{Criteria\}$\\
Instructions:\\ * Score each summary based on the above criteria.\\ * Calculate an overall score for each summary as the sum of all criteria scores (maximum 50).\\ * Conclude by identifying which summary is better overall.\\ * Calculate a score ratio of Summary 1 to Summary 2 (Summary 1 Score $\div$ Summary 2 Score).\\
Output Format: $\{Format\}$
\end{template}
\noindent where we force \exper{Llm}s to score the responses of the adjacent steps generated by itself. By extracting the score ratio from $\{Format\}$, we can estimate the fraction of the target distribution.

For instruction following tasks, we use the pairwise template derived from the Alpaca-Eval, which emphasizes that the length of the generated content and the position of the presentation should not be a bias in scoring.
\begin{template}{$\Phi(y_i,y_{i-1}|x)$}
\texttt{[SYSTEM]:} You are a highly efficient assistant, who evaluates and selects the best large language model (LLMs) based on the quality of their responses to a given instruction.\\
This process will be used to create a leaderboard reflecting the most accurate and human-preferred answers.\vspace{3pt}\\
\texttt{[USER]:} I require a leaderboard for various large language models.
I'll provide you with an instruction given to these models and their corresponding responses.
Your task is to assess these responses, provide a quantitative assessment of their performance based on the criteria below, and select the model that produces the best output from a human perspective.
Avoid any position biases and ensure that the order in which the responses were presented does not influence your decision.\\
Instruction: $\{x\}$\\
Response 1: $\{y_{i}\}$\\
Response 2: $\{y_{i-1}\}$\\
Tasks:\\
* Score each response based on the above criteria.\\
* Calculate an overall score for each response as the sum of all criteria scores (maximum 60).\\
* Conclude by identifying which response is better overall.\\
* Calculate a score ratio of Response 1 to Response 2 (Response 1 Score ÷ Response 2 Score).\\
Output Format: $\{Format\}$
\end{template}

\subsection{Propose New States}

\paragraph{Proposal Distribution} For the abstractive summarization task, the prompt template for sampling from the proposal distribution $p(y_i|y_{i \minus 1},x)$ is: 
\begin{template}{$y_i\sim p(y_i|y_{i \minus 1},x)$}
\texttt{[SYSTEM]:} You are a powerful abstractive summarizer. \vspace{3pt}\\
\texttt{[USER]:} Document: $\{x\}$\\
Based on the previous document, provide a high-quality summary in exactly $\{\ell\}$ words:\vspace{3pt}\\
\texttt{[ASSISTANT]:} Summary: $\{y_{i-1}\}$
\tcblower
\texttt{[USER]:} Please generate a new summary based on the previous one: \vspace{3pt}\\
\texttt{[ASSISTANT]:} Summary: $\{y_{i}\}$
\end{template}
\noindent The template for instruction following task is:
\begin{template}{$y_i\sim p(y_i|y_{i \minus 1},x)$}
\texttt{[USER]:} Answer the following instruction using $\{\ell\}$ words or less. \texttt{\textbackslash n\textbackslash n} $\{x\}$ \vspace{3pt}\\
\texttt{[ASSISTANT]:} Answer:\texttt{\textbackslash n}$\{y_{i-1}\}$
\tcblower
\texttt{[USER]:} Please generate a new answer based on the previous one: \vspace{3pt}\\
\texttt{[ASSISTANT]:} Answer: \texttt{\textbackslash n} $\{y_{i}\}$
\end{template}

\paragraph{Importance Distribution} We split the importance distribution into two segments. When the candidate length is far from the target length $\mathscr{D}(y,\ell)>3$, we use a looser objective so that \exper{Llm}s can have more opportunities for semantic organization, which is beneficial for the quality of generation.
The template for abstractive summarization is:
\begin{template}{$y_i\sim q(y_i|y_{i \minus 1},x)$}
\texttt{[SYSTEM]:} You are a powerful abstractive summarizer. \vspace{3pt}\\
\texttt{[USER]:} Document: \texttt{\textbackslash n} $\{x\}$ \texttt{\textbackslash n\textbackslash n} Based on the previous document, provide a high-quality summary in exactly $\{\ell\}$ words: \vspace{3pt}\\
\texttt{[ASSISTANT]:} Summary: \texttt{\textbackslash n} $\{y_{i-1}\}$
\tcblower
\texttt{[USER]:} The generated summary is too (long / short) at $\{\mathtt{Len}(y)\}$ words.\\
Please improve it to be exactly $\{\ell\}$ words by (focusing on the core ideas and removing some redundant details / adding some details and maintaining clarity and relevance): \vspace{3pt}\\
\texttt{[ASSISTANT]:} Summary: \texttt{\textbackslash n} $\{y_{i}\}$
\end{template}
\noindent The prompt template for instruction following is:
\begin{template}{$y_i\sim q(y_i|y_{i \minus 1},x)$}
\texttt{[USER]:} Answer the following instruction using $\{\ell\}$ words or less. \texttt{\textbackslash n\textbackslash n} $\{x\}$ \vspace{3pt}\\
\texttt{[ASSISTANT]:} Answer: \texttt{\textbackslash n} $\{y_{i-1}\}$
\tcblower
\texttt{[USER]:} The generated answer is too long at $\{\mathtt{Len}(y)\}$ words.
Please improve it to be exactly $\{\ell\}$ words or less by focusing on the core contents and removing any unhelpful, irrelevant, or inaccurate parts:\vspace{3pt}\\
\texttt{[ASSISTANT]:} Answer: \texttt{\textbackslash n} $\{y_{i}\}$
\end{template}
\noindent When the candidate length is close to the target length $\mathscr{D}(y,\ell)\leq 3$, we force an accurate length control such that \exper{Llm}s are required to add or delete an exact number of words. The prompt template for abstractive summarization is:
\begin{template}{$y_i\sim q(y_i|y_{i \minus 1},x)$}
\texttt{[SYSTEM]:} You are a powerful abstractive summarizer. \vspace{3pt}\\
\texttt{[USER]:} Document: \texttt{\textbackslash n} $\{x\}$ \texttt{\textbackslash n\textbackslash n} Based on the previous document, provide a high-quality summary in exactly $\{\ell\}$ words: \vspace{3pt}\\
\texttt{[ASSISTANT]:} Summary: \texttt{\textbackslash n} $\{y_{i-1}\}$
\tcblower
\texttt{[USER]:} Please (delete / add) $\{\mathscr{D}(y,\ell)\}$ words appropriately based on the previous summary: \vspace{3pt}\\
\texttt{[ASSISTANT]:} Summary: \texttt{\textbackslash n} $\{y_{i}\}$
\end{template}
\noindent The prompt template for instruction following is:
\begin{template}{$y_i\sim q(y_i|y_{i \minus 1},x)$}
\texttt{[USER]:} Answer the following instruction using $\{\ell\}$ words or less. \texttt{\textbackslash n\textbackslash n} $\{x\}$ \vspace{3pt}\\
\texttt{[ASSISTANT]:} Answer: \texttt{\textbackslash n} $\{y_{i-1}\}$
\tcblower
\texttt{[USER]:} The generated answer is too long at $\{\mathtt{Len}(y)\}$ words.
Please delete $\{\mathscr{D}(y,\ell)\}$ words appropriately based on the previous response:\vspace{3pt}\\
\texttt{[ASSISTANT]:} Answer: \texttt{\textbackslash n} $\{y_{i}\}$
\end{template}

\begin{table}[t]
  \centering
  \begin{tabular}{l||c|c|c|c}
    \hline
    \textbf{Models} & \textbf{Top-K} & \textbf{Top-P} & \textbf{Temp.} & \textbf{Rep.}\\
    \hline
    
    \hline
    \hline
    \exper{Qwen2.5}& $20$ & $0.8$ & $0.7$ & $1.05$ \\
    \exper{Llama2}& $50$ & $0.9$ & $0.6$ & $1.00$ \\
    \exper{Llama3}& $50$ & $0.9$ & $0.6$ & $1.00$ \\
    \exper{Llama3.1}& $50$ & $0.9$ & $0.6$ & $1.00$ \\
    \hline
    \hline

    \hline
  \end{tabular}
  \caption{Generation configurations of \exper{Llm}s.}
  \label{apptab:config}
\end{table}
\section{Experimental Details}
Our experiments are implemented on the \textit{Huggingface Transformers} package\footnote{\url{https://github.com/huggingface/transformers}}. All \exper{Llm}s we used are the chat version trained with supervised fine tuning, where \exper{Llama2} and \exper{Qwen2.5} have $7$B parameters while \exper{Llama3} and \exper{Llama3.1} have $8$B parameters. The generation configurations of each model are set by default, as demonstrated in \Cref{apptab:config}. There is no training stage of our framework, and the inference is performed on an NVIDIA A100 80GB GPU with a random seed of $0$. 

For the CNN/Daily Mail dataset, we randomly choose $1000$ samples from the $3.0$ version of the test set, since the instruction following task contains $1042$ samples ($802$ from Alpaca-Eval-LI and $240$ from MT-Bench-LI).

\section{Analyses}
\begin{table}[t]
  \setlength{\tabcolsep}{4.5pt}
  \centering
  \begin{tabular}{c||c|c|c||c}
    \hline
    \textbf{Trials} &\textbf{\exper{Acc}$\uparrow$} & \textbf{\exper{L1}$\downarrow$} & \textbf{\exper{L2}$\downarrow$} & \textbf{\exper{Rouge-(1/2/L)}} \\
    \hline
    
    \hline
    \hline
    \multicolumn{5}{l}{\textbf{\exper{Llama3}}}\\
    \hline
    $0$ & $\phantom{0}9.1\%$ & $\phantom{}4.78$ & $\phantom{}6.10$ & $0.39/0.14/0.25$\\
    $1$ &$\phantom{}48.3\%$ & $\phantom{}0.98$ & $\phantom{}1.80$ & $0.38/0.14/0.24$\\
    $2$ &$\phantom{}64.8\%$ & $\phantom{}0.58$ & $\phantom{}1.05$ & $0.38/0.13/0.24$\\
    $3$ &$\phantom{}68.4\%$ & $\phantom{}0.44$ & $\phantom{}0.85$ & $0.38/0.13/0.24$\\
    $4$ &$\phantom{}72.3\%$ & $\phantom{}0.36$ & $\phantom{}0.75$ & $0.38/0.14/0.25$\\
    $5$ &$\phantom{}78.6\%$ & $\phantom{}0.29$ & $\phantom{}0.66$ & $0.38/0.14/0.25$\\
    \hline
    \hline

    \hline
  \end{tabular}
  \caption{Analysis of the iteration trial on the \exper{CnnDm} dataset, where the beam size is $16$.}
  \label{apptab:trial}
\end{table}

\begin{table}[t]
  \setlength{\tabcolsep}{3.8pt}
  \centering
  \begin{tabular}{c||c|c|c||c}
    \hline
    \textbf{Beams} & \textbf{\exper{Acc}$\uparrow$} & \textbf{\exper{L1}$\downarrow$} & \textbf{\exper{L2}$\downarrow$} & \textbf{\exper{Rouge-(1/2/L)}}\\
    \hline
    
    \hline
    \hline
    \multicolumn{5}{l}{\textbf{\exper{Llama3}}}\\
    \hline
    $\phantom{0}0$ & $\phantom{0}9.1\%$ & $\phantom{}4.78$ & $\phantom{}6.10$ & $0.39/0.14/0.25$\\
    $\phantom{0}1$ & $\phantom{}36.4\%$ & $\phantom{}1.63$ & $\phantom{}2.51$ & $0.39/0.14/0.25$\\
    $\phantom{0}2$ & $\phantom{}47.2\%$ & $\phantom{}1.08$ & $\phantom{}2.10$ & $0.38/0.14/0.25$\\
    $\phantom{0}4$ & $\phantom{}55.8\%$ & $\phantom{}0.72$ & $\phantom{}1.25$ & $0.38/0.14/0.25$\\
    $\phantom{0}8$ & $\phantom{}67.2\%$ & $\phantom{}0.47$ & $\phantom{}0.93$ & $0.38/0.14/0.25$\\
    $16$ & $\phantom{}78.6\%$ & $\phantom{}0.29$ & $\phantom{}0.66$ & $0.38/0.14/0.25$\\
    \hline
    \hline

    \hline
  \end{tabular}
  \caption{Analysis of the beam size on the \exper{CnnDm} dataset, where the iteration trial is $5$.}
  \label{apptab:beam}
\end{table}

\begin{table}[t]
  \setlength{\tabcolsep}{3pt}
  \centering
  \begin{tabular}{l||c|c|c||c}
    \hline
    \textbf{Samp.} & \textbf{\exper{Acc}$\uparrow$} & \textbf{\exper{L1}$\downarrow$} & \textbf{\exper{L2}$\downarrow$} & \textbf{\exper{Rouge-(1/2/L)}} \\
    \hline
    
    \hline
    \hline
    \exper{Inst} & $\phantom{0}3.2\%$ & $8.64$ & $10.83$ & $0.33/0.10/0.21$\\
    \exper{Rand} & $17.7\%$ & $3.54$ & $\phantom{0}5.23$ & $0.33/0.10/0.21$\\
    \exper{Mh} & $14.1\%$ & $4.73$ & $\phantom{0}7.65$ & $0.32/0.10/0.21$\\
    \exper{Mh+Is} & $24.6\%$ & $2.41$ & $\phantom{0}4.02$ & $0.33/0.10/0.21$\\
    \hline
    \hline

    \hline
  \end{tabular}
  \caption{Ablation study of \exper{Qwen2.5} on \exper{CnnDm}, where the iteration trial is $5$ and the beam size is $1$.}
  \label{apptab:ablation_qwen2.5}
\end{table}

We demonstrate the hyperparameter analyses in \Cref{apptab:trial,apptab:beam}. Similar to the observation in \S \ref{sec:analyze}, the marginal effect of \exper{Llama3} as the sampling space grows is between \exper{Qwen2.5} and \exper{Llama3.1}.

For ablation studies in \exper{Qwen2.5} (\Cref{apptab:ablation_qwen2.5}) and \exper{Llama3} (\Cref{apptab:ablation_llama3}), our sampling framework outperforms other methods. However, since the instruction following capabilities of these models are not as powerful as \exper{Llama3.1}, their improvement may not be as significant.

\begin{table}[H]
  \setlength{\tabcolsep}{3pt}
  \centering
  \begin{tabular}{l||c|c|c||c}
    \hline
    \textbf{Samp.} & \textbf{\exper{Acc}$\uparrow$} & \textbf{\exper{L1}$\downarrow$} & \textbf{\exper{L2}$\downarrow$} & \textbf{\exper{Rouge-(1/2/L)}} \\
    \hline
    
    \hline
    \hline
    \exper{Inst} & $\phantom{0}9.1\%$ & $4.78$ & $\phantom{0}6.10$ & $0.39/0.14/0.25$\\
    \exper{Rand} & $30.7\%$ & $3.05$ & $\phantom{0}5.49$ & $0.38/0.14/0.25$\\
    \exper{Mh} & $30.1\%$ & $4.07$ & $\phantom{0}5.98$ & $0.38/0.13/0.24$\\
    \exper{Mh+Is} & $36.4\%$ & $1.63$ & $\phantom{0}2.51$ & $0.39/0.14/0.25$\\
    \hline
    \hline

    \hline
  \end{tabular}
  \caption{Ablation study of \exper{Llama3} on \exper{CnnDm}, where the iteration trial is $5$ and the beam size is $1$.}
  \label{apptab:ablation_llama3}
\end{table}
\section{Generated Cases}
The generated cases of \exper{Llama3.1} on the Alpaca-Eval-LI dataset are demonstrated in \Cref{apptab:case}.
\begin{table*}
    \centering
    \begin{tabular}{p{8.1cm} p{7cm}}
        \toprule
        \textbf{\exper{Llama3.1}} \exper{Inst} & \textbf{\exper{Llama3.1}} \exper{Ours} \\
        \midrule
        \midrule
        \multicolumn{2}{l}{\parbox{15cm}{Answer the following instruction using 46 words or less.\\Is the US border open to Canada?}}\\
        \midrule
        \textit{Number of Words:} $48$ & \textit{Number of Words:} $44$ \\
        \specialrule{0pt}{2pt}{2pt}
        The US border with Canada is open, but with restrictions. Land and sea borders are open for essential travel, while air travel requires a valid visa or electronic travel authorization. Some restrictions may apply, so it's best to check the official government websites for the latest information. & The US-Canada border is open, but with restrictions. Travelers must provide proof of COVID-19 vaccination, a negative test, or recovery from the virus. Additionally, some land ports of entry may be closed or have limited hours, so it's best to check before traveling.\\
        \midrule
        \midrule
        \multicolumn{2}{l}{\parbox{15cm}{Answer the following instruction using 50 words or less.\\Write me a top 10 list of the funniest ways to die.}}\\
        \midrule
        \textit{Number of Words:} $110$ & \textit{Number of Words:} $48$ \\
        \specialrule{0pt}{2pt}{2pt}
        \parbox[t]{8.1cm}{Please note that this is a dark humor list and not meant to be taken seriously. Here's a list of the funniest ways to die (in a fictional context):\\1. Dying from laughing too hard at a dad joke.\\2. Choking on a chicken nugget.\\3. Falling into a giant pile of pillows.\\4. Getting hit by a runaway inflatable unicorn.\\5. Accidentally hugging a cactus.\\6. Dying from a massive sneeze.\\7. Getting stuck in a giant toilet.\\8. Falling off a unicycle while riding a unicycle.\\9. Dying from eating too much wasabi.\\10. Being attacked by a swarm of bees while trying to get a selfie.} & \parbox[t]{7cm}{Here's a list of humorous ways to die:\\1. Laughing too hard.\\2. Choking on a whoopee.\\3. Drowning in Jell-O.\\4. Falling into feathers.\\5. Being crushed by cats.\\6. Overdosing on laughter.\\7. Sneezing from feather.\\8. Excessive flatulence.\\9. Cheese coma.\\10. Snoring loudly.}\\

        \midrule
        \midrule
        \multicolumn{2}{l}{\parbox{15cm}{Answer the following instruction using 128 words or less.\\Who made Berlin?}}\\
        \midrule
        \textit{Number of Words:} $151$ & \textit{Number of Words:} $125$ \\
        \specialrule{0pt}{2pt}{2pt}
        \parbox[t]{8.1cm}{Berlin, the capital city of Germany, has a complex and multifaceted history. The city's origins date back to the 13th century when it was a small trading post on the Spree River. Over time, Berlin grew and became a significant center for trade, culture, and politics.\\In 1237, the Margraviate of Brandenburg was founded, and Berlin became its capital. In 1415, the Hohenzollern dynasty took control of the city, and Berlin continued to grow and prosper. Throughout the centuries, Berlin was influenced by various cultures and empires, including the Prussian Empire, the German Empire, and the Weimar Republic.\\The city's modern history is marked by significant events, including World War I and II, the division of Germany during the Cold War, and the fall of the Berlin Wall in 1989. Today, Berlin is a vibrant and diverse city, known for its rich history, cultural attractions, and thriving arts scene.} & \parbox[t]{7cm}{Berlin, the capital of Germany, has a complex and evolving history. The area has been inhabited since the Stone Age, with various tribes and cultures contributing to its development.\\The city of Berlin as we know it today began to take shape in the 13th century, when it was a small trading post on the Spree River. In 1237, the Margraviate of Brandenburg was established, and Berlin became the capital. Over the centuries, the city grew and was influenced by various rulers, including the Hohenzollern dynasty, which played a significant role in shaping modern Berlin.\\In 1871, Berlin became the capital of Germany, and it continued to grow and evolve throughout the 20th century, experiencing significant changes under the Weimar regime, and the post-war period.}\\

        \bottomrule
    \end{tabular}
    \caption{Generated cases of \exper{Llama3.1} on the Alpaca-Eval-LI dataset.}
    \label{apptab:case}
\end{table*}
\end{document}